\documentclass[11pt]{article}
\usepackage[margin=1in]{geometry}
\usepackage{amsmath}
\usepackage{amssymb}
\usepackage{booktabs}
\usepackage{hyperref}
\usepackage{natbib}
\usepackage{graphicx}
\usepackage{authblk}

\title{Beyond Aggregate Calibration: Decomposing Income-Conditional Recall Disparities in Automated Credit Default Prediction}
\author[1]{Sai Srikar Boddupalli}
\affil[1]{Independent Researcher}
\date{}

\begin{document}
\maketitle

\begin{abstract}
Data-centric curation pipelines frequently rely on model confidence scores to flag and filter noisy or mislabeled training instances. Evaluating this filtering convention on a large-scale consumer lending sample (LendingClub, $N = 1{,}344{,}936$) uncovers an underlying demographic asymmetry: high-income defaulters are disproportionately classified as label noise relative to low-income defaulters (Cram\'er's $V \approx 0.03$--$0.07$). Re-examining this behavior through the lens of equal opportunity \citep{hardt2016} reveals a far more severe discrepancy: a 16.86 percentage point gap in true positive rate (recall) between high- and low-income borrowers who ultimately defaulted. Implementing a sequential feature-blinding methodology allows us to isolate the drivers of this disparity across three distinct mechanisms: (1) direct reliance on self-reported applicant income; (2) algorithmic absorption of upstream institutional bias encoded within origination interest rates; and (3) a residual disparity (3.55 percentage points in cross-validation; 2.56 percentage points on a held-out test partition, $Z = -4.04$, $p < 0.0001$) that remains even after purging both income and interest rates from the model. Out-of-sample signed SHAP valuations demonstrate that this residual gap is maintained by structural proxies, most notably loan amount and home ownership status. These empirical findings show that simply blinding an algorithm to sensitive attributes fails to ensure fairness when institutional pricing decisions and behavioral proxy variables collectively reconstruct the omitted signals. We outline the practical implications of these findings for auditing data-centric AI workflows within regulated financial institutions.
\end{abstract}

\section{Introduction}

Automated machine learning systems increasingly support or replace human judgment in high-stakes consumer credit underwriting. Prior literature confirms that removing protected demographic attributes from training datasets rarely prevents models from reproducing structural disparities, as high-capacity algorithms readily reconstruct sensitive characteristics from correlated features \citep{fuster2022, bartlett2022}. In a separate line of inquiry, the data-centric AI community has developed automated techniques for dataset hygiene, often using a classifier's prediction confidence to detect and discard suspected label noise. Despite their shared prominence, the intersection between credit scoring fairness and confidence-based dataset pruning remains largely unexplored.

This investigation begins with a diagnostic observation: applying standard confidence-threshold filtering to a consumer credit dataset removes ``noisy'' default labels at significantly different rates across borrower income brackets. We leverage this initial asymmetry to investigate a deeper structural question: when an algorithm's predicted probabilities contradict ground-truth default events, does the likelihood of treating a failure as an anomaly skew systematically along socioeconomic lines? We show that such an income-dependent bias exists and execute a sequential feature-blinding protocol to explain its mechanical drivers.

Our primary contributions are fourfold:

\begin{itemize}
    \item \textbf{Empirical Audit of Label Cleaning:} We demonstrate that confidence-based label filtering introduces a statistically significant, income-conditional skew within a real-world consumer lending dataset.
    \item \textbf{Fairness Metric Reframing:} We evaluate model behavior using the equal opportunity criterion, uncovering a substantial recall deficit for minority-class outcomes among high-income borrowers.
    \item \textbf{Mechanism Decomposition:} Using sequential feature blinding and out-of-sample signed SHAP analysis, we trace the disparity to direct income utilization, inherited institutional interest-rate pricing, and downstream proxy variables including loan amount and asset ownership.
    \item \textbf{Methodological Verification:} We confirm our results across distinct model families (gradient-boosted trees and logistic regression) and validate generalization using a strict held-out test partition, while explicitly documenting data limitations such as anonymized repeat-borrower records.
\end{itemize}

\section{Related Work}

\textbf{Fairness Metrics in Supervised Learning.} \citet{hardt2016} proposed equalized odds and equal opportunity as alternatives to demographic parity, requiring algorithms to achieve equal true positive rates across distinct demographic segments. We apply equal opportunity as our core evaluation standard because it directly quantifies how equitably a classifier predicts minority-class outcomes (actual defaults) within each group, avoiding aggregate accuracy measures that obscure subgroup-level failures.

\textbf{Imperfect Group Information.} \citet{awasthi2020} showed that equalized-odds post-processing interventions degrade rapidly when auxiliary demographic attributes are noisy or imperfectly captured. Their work illustrates the inherent vulnerability of fairness adjustments to imperfect feature representations, reinforcing the necessity of validating proxy mechanisms on unseen out-of-sample data.

\textbf{Label Noise and Fairness.} \citet{wu2022} formulated a theoretical framework for instance-dependent label noise, proving mathematically that feature-correlated noise disproportionately distorts group-conditional performance. Our observation that confidence-based filtering prunes ground-truth defaults at varying rates across income tiers provides empirical confirmation of the feature-correlated noise dynamics modeled in their framework.

\textbf{Measuring and Defining Fairness.} Following \citet{corbettdavies2023}, who analyze the mathematical tradeoffs among competing fairness definitions, we report both absolute and relative disparity measures throughout this study to ensure an accurate representation of practical impact.

\textbf{Machine Learning and Credit Market Inequality.} \citet{fuster2022} demonstrated that flexible machine learning architectures in mortgage underwriting triangulate excluded racial and gender characteristics via auxiliary variables. Similarly, \citet{bartlett2022} documented persistent pricing disparities in algorithmic lending models despite formal attribute blinding. We contribute to this literature by analyzing an open consumer lending sample and separating the observed disparity into an upstream institutional component (origination interest rate) and a downstream borrower-selection component (loan amount).

\section{Data and Methods}

\subsection{Dataset}

We analyze the publicly accessible LendingClub accepted-loans dataset covering origination years 2007 through the fourth quarter of 2018. Limiting our scope to loans with terminal outcomes (``Fully Paid'' or ``Charged Off'') yields a final analytical sample of $N = 1{,}344{,}936$ records. We define the binary target variable \texttt{is\_default} $= 1$ for charged-off accounts. Borrowers are partitioned into three post-hoc demographic brackets based on self-reported annual income at origination: Low ($<\$50{,}000$), Middle (\$50{,}000--\$100{,}000), and High ($>\$100{,}000$). These demographic groupings are utilized exclusively for post-hoc evaluation and are withheld from the feature space during blinded training executions.

\subsection{Feature Set and Preprocessing}

We evaluate algorithmic performance across two distinct feature configurations:

\begin{itemize}
    \item \textbf{Minimal Baseline (4 features):} \texttt{loan\_amnt}, \texttt{annual\_inc}, \texttt{int\_rate}, and \texttt{dti}.
    \item \textbf{Expanded Feature Set (11 base features):} Adds credit profile indicators (\texttt{fico\_range\_low}, \texttt{revol\_util}, \texttt{pub\_rec\_bankruptcies}), housing and employment metrics (\texttt{emp\_length}, \texttt{home\_ownership}), and contract specifications (\texttt{purpose}, \texttt{term}).
\end{itemize}

Missing data is preserved as an informative signal rather than dropped or mean-imputed. Numeric columns exhibiting structural missingness are filled with designated sentinel values (0 or $-1$) and accompanied by a binary \texttt{\_missing} indicator variable. All preprocessing steps are encapsulated within a scikit-learn \texttt{ColumnTransformer} pipeline to prevent information leakage across cross-validation and holdout boundaries.

\subsection{Model Training}

We train gradient-boosted decision trees using XGBoost, setting the \texttt{scale\_pos\_weight} hyperparameter to the negative-to-positive class ratio of the training sample ($\approx 4$:1) to account for class imbalance. Models are evaluated using 5-fold cross-validation stratified jointly on default status and income bracket. Headline metrics are subsequently verified against an untouched 80/20 held-out test split separated prior to any pipeline transformations. We also implement a regularized logistic regression baseline with standardized numerical inputs to check for model architecture dependence.

\subsection{Sequential Feature-Blinding Design}

To isolate the mechanical drivers of income-conditional recall gaps, we evaluate three progressive model specifications:

\begin{enumerate}
    \item \textbf{Full Model:} Utilizes the entire feature set, explicitly including \texttt{annual\_inc}.
    \item \textbf{Income-Blind Model:} Excludes \texttt{annual\_inc} from the training features.
    \item \textbf{Double-Blind Model:} Excludes both \texttt{annual\_inc} and \texttt{int\_rate}.
\end{enumerate}

We omit \texttt{int\_rate} in the third specification because interest rates are assigned by human underwriters at origination; retaining this variable allows the algorithm to exploit an upstream institutional risk appraisal that already correlates strongly with borrower income.

\subsection{Metrics and Statistical Testing}

We adopt the equal opportunity criterion \citep{hardt2016} as our primary fairness metric, evaluating group-conditional true positive rates (TPR) exclusively among actual defaulters. Aggregate calibration measures can obscure systemic misclassification concentrated within minority true-default subpopulations. Statistical significance is evaluated using two-proportion $z$-tests and chi-square tests of independence, reporting both $p$-values and effect sizes (Cram\'er's $V$).

\subsection{Mechanism Analysis (SHAP)}

We compute SHAP (SHapley Additive exPlanations) values via \texttt{TreeExplainer} on the XGBoost models to determine feature-level contributions to individual predictions. To prevent in-sample memorization from confounding structural feature attribution, SHAP values are extracted strictly from the 20\% held-out test partition using models trained solely on the complementary 80\% training set.

\section{Results}

\subsection{Motivating Observation: Confidence-Based Cleaning Discards Labels Unevenly}

Using the minimal baseline model, we flag ``noisy'' instances using a probability threshold: actual defaulters (\texttt{is\_default} $= 1$) whose out-of-fold predicted default probability drops below 0.15 are identified as potential label noise. Applying this filtering convention uncovers a notable demographic imbalance:

\begin{table}[h]
\centering
\begin{tabular}{lrrr}
\toprule
Income Bracket & Total True Defaults & Discarded as Noise & Discard Rate \\
\midrule
High & 44{,}069 & 1{,}057 & 2.40\% \\
Middle & 135{,}197 & 1{,}345 & 0.99\% \\
Low & 89{,}293 & 339 & 0.38\% \\
\bottomrule
\end{tabular}
\caption{Confidence-threshold discard rates by income bracket.}
\label{tab:discard-rates}
\end{table}

A chi-square test confirms that this demographic disparity is statistically significant ($\chi^2 = 1192.26$, $p < 0.0001$). Although the absolute effect size is small (Cram\'er's $V = 0.0666$, falling within the $\approx 0.03$--$0.07$ range observed across tested thresholds), the measurement demonstrates that standard confidence heuristics introduce an income-conditional skew, resulting in a 2.02 percentage-point absolute gap between high- and low-income discard rates, or roughly 6x in relative terms.

\subsection{Reframing with Equal Opportunity Reveals a Larger Disparity}

Evaluating model performance exclusively through discard rates understates the practical severity of the disparity by conflating sample filtering with prediction failure. Assessing the full expanded-feature model under equal opportunity measures the group-conditional TPR directly among actual defaulters. The model achieves 72.84\% recall for low-income defaults and 66.69\% for middle-income defaults, but drops to 55.98\% for high-income defaults. This yields a raw 16.86 percentage point recall gap between the highest- and lowest-earning tiers, demonstrating that aggregate calibration masks severe misclassification within the high-income default class.

\begin{figure}[h]
\centering
\includegraphics[width=0.8\textwidth]{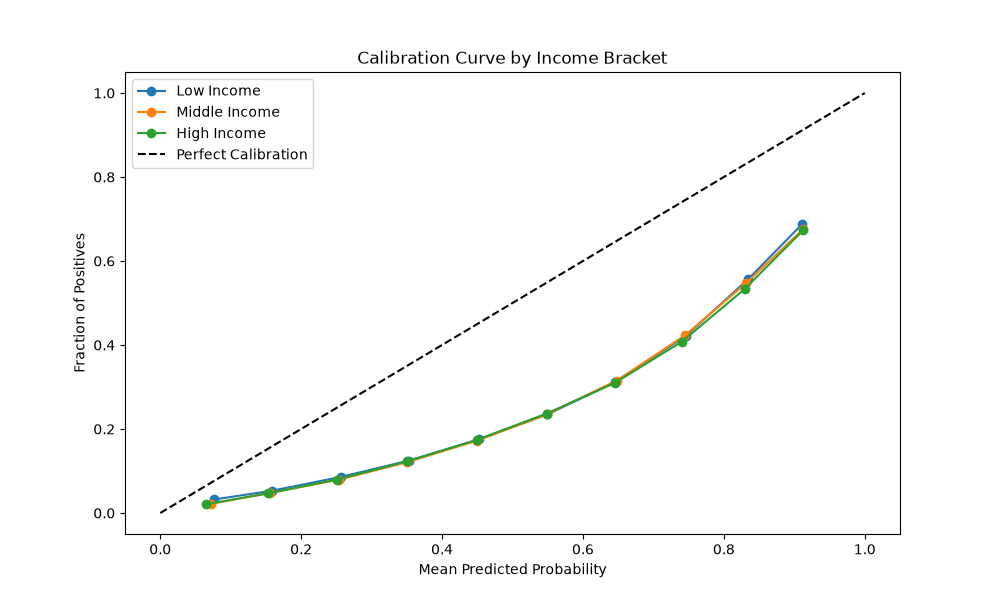}
\caption{Calibration curve by income bracket, showing near-identical aggregate calibration despite the underlying recall disparity.}
\label{fig:calibration}
\end{figure}

\subsection{Decomposition via Sequential Blinding}

Removing \texttt{annual\_inc} from the feature space (the income-blind variant) compresses the High-Low TPR gap from 16.86 percentage points to 7.49 percentage points (High $= 61.35\%$, Middle $= 67.50\%$, Low $= 68.84\%$). Approximately half of the initial recall disparity is mechanically driven by the model's direct access to borrower income.

Removing \texttt{int\_rate} (the double-blind variant) further compresses the gap to 3.55 percentage points in cross-validation (High $= 59.23\%$, Middle $= 63.70\%$, Low $= 62.78\%$) and 2.56 percentage points on the held-out test partition ($Z = -4.04$, $p < 0.0001$, 95\% CI $[1.31\%, 3.80\%]$). Because the double-blind model retains meaningful discriminative capacity (holdout AUC $\approx 0.69$), the residual disparity reflects a genuine structural effect rather than model collapse.

\paragraph{Note on Ordering.} The strict monotonic ordering (High $<$ Middle $<$ Low) observed in our cross-validated estimates does not hold precisely in the double-blind model, where middle-income recall (63.70\%) slightly exceeds low-income recall (62.78\%). Section~4.5 confirms that this anomaly is a composition artifact rather than a reversal of the underlying proxy mechanism.

\subsection{Mechanism: Interest Rate as an Inherited Upstream Signal}

Out-of-sample signed SHAP analysis on high-income false negatives within the income-blind model (actual defaulters incorrectly classified as safe on unseen test data) identifies \texttt{int\_rate} as the dominant variable suppressing default risk predictions. The mean SHAP contribution for interest rate is $-0.355$, exceeding the magnitude of the second-largest feature (\texttt{dti} at $-0.092$) by nearly fourfold.

An examination of raw distributions explains the underlying mechanics: high-income true defaulters were assigned lower origination interest rates (median 14.49\%) than low-income true defaulters (median 15.31\%). Human underwriters priced an expectation of safety into high-income applications at origination; when denied direct income access, the classifier absorbs that human assessment by treating low interest rates as an indicator of creditworthiness.

\begin{figure}[h]
\centering
\includegraphics[width=0.8\textwidth]{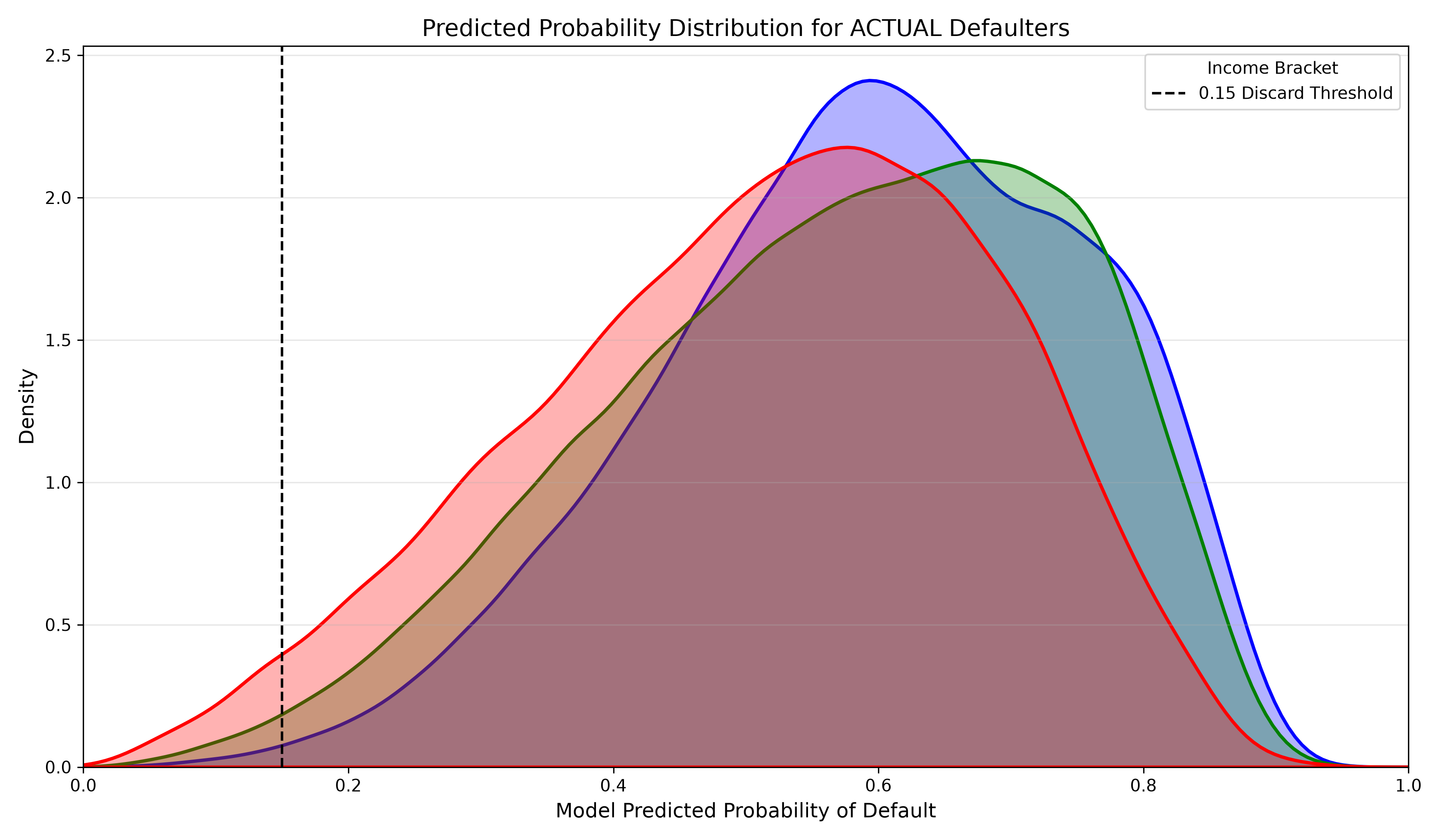}
\caption{Predicted probability distribution for actual defaulters, by income bracket. The dashed line marks the 0.15 discard threshold used in Section 4.1.}
\label{fig:kde}
\end{figure}

\subsection{Residual Mechanism: Loan Amount and Home Ownership as Proxies}

Even when explicitly blinded to both income and interest rate, a statistically significant recall gap (2.56--3.55 percentage points) persists. Double-blind SHAP analysis comparing middle- and low-income false negatives identifies \texttt{loan\_amnt} as the dominant driver of this remaining gap (Middle mean SHAP $\approx -0.041$, Low mean SHAP $\approx -0.106$; absolute difference $\approx 0.065$), followed by \texttt{home\_ownership\_MORTGAGE}. We verify loan amount's role as an income proxy via a Spearman rank correlation between \texttt{annual\_inc} and \texttt{loan\_amnt} of 0.481, which represents the strongest pairwise relationship among all continuous variables analyzed.

This proxy mechanism also resolves the non-monotonic ordering noted in Section~4.3 (Middle $>$ Low). When true positive rates are evaluated across isolated loan-amount deciles rather than in aggregate, high-income borrowers exhibit the lowest recall across nearly every decile, while middle-income recall remains below low-income recall in lower deciles. However, because middle-income defaulters borrow larger absolute amounts (median \$16{,}000) than low-income defaulters (median \$10{,}000), and higher loan tiers carry higher predicted default probabilities across all groups, the aggregate recall for middle-income borrowers is shifted upward by their underlying loan-size distribution.

\subsection{Robustness Checks}

\begin{itemize}
    \item \textbf{Validation Schemes:} Core empirical findings---including the income-blind TPR gap, the double-blind TPR gap, and individual SHAP valuations---replicate within approximately 1 percentage point between 5-fold cross-validation and the untouched 20\% holdout sample.
    \item \textbf{Model Architecture:} Implementing a regularized logistic regression pipeline on the double-blind feature set reproduces the observed recall disparity (High $= 57.74\%$, Middle $= 62.82\%$, Low $= 61.38\%$), confirming that the proxy effect is embedded in the data structure rather than arising as an artifact of decision-tree partitioning.
\end{itemize}

\section{Discussion and Limitations}

Our empirical decomposition demonstrates that naive fairness interventions---such as removing a protected attribute from a training dataset---fail to eliminate demographic disparities in consumer lending. A significant portion of the original bias is reintroduced through upstream institutional variables (interest rates) that encode prior human judgment, while a smaller residual gap persists via downstream behavioral proxies (loan size and asset ownership).

\paragraph{Limitations.} We cannot exhaustively rule out additional non-linear proxy relationships beyond loan amount and home ownership; our 2.56 percentage point residual estimate represents a lower bound on remaining proxy effects. Furthermore, because the issuing platform fully redacted the borrower-level identifier (\texttt{member\_id}), we cannot test for repeat-borrower data leakage across validation splits. We document this as an inherent constraint of the dataset artifact. Finally, the Section~4.1 discard-rate percentages are sensitive to the chosen confidence threshold, and we present them as a motivating observation rather than our primary empirical claim.

\paragraph{Scope.} These findings apply specifically to income-conditional recall disparities within U.S.-based unsecured consumer credit between 2007 and 2018. We make no claim that these specific proxies (interest rate and loan size) generalize identically across distinct financial products like mortgages, or across protected demographic classes such as race and gender, which were unavailable in this public sample.

\section{Conclusion}

This study demonstrates that an open consumer lending dataset contains income-conditional disparities that impact both data-cleaning heuristics and minority-class recall. Combining sequential feature blinding with out-of-sample SHAP analysis allowed us to decompose an aggregate recall disparity into direct feature use, inherited institutional bias, and structural proxy mapping. For data-centric AI workflows deployed in regulated financial environments, fairness audits must extend beyond explicit attributes to account for upstream pricing decisions and downstream transaction geometries that systematically reconstruct omitted variables.

\bibliographystyle{plainnat}
\bibliography{references}

\appendix
\section{Reproducibility Notes}

\subsection{Execution Pipeline and Script Inventory}

The empirical results, statistical tests, and mechanism decompositions reported in this study were generated across an iterative 17-script experimental pipeline (\texttt{01\_inspect\_data.py} through \texttt{17\_final\_methods\_checks.py}). Key mappings:

\begin{itemize}
    \item \texttt{01\_inspect\_data.py}--\texttt{06\_fairness\_audit.py}: Data ingestion, preprocessing, and the exploratory confidence-threshold audit establishing the motivating observation in Section~4.1 (final statistics logged in \texttt{17\_final\_methods\_checks.py}).
    \item \texttt{07\_robust\_baseline\_fixed.py}--\texttt{09\_blind\_model\_audit.py}: Full, income-blind, and double-blind XGBoost training via stratified 5-fold cross-validation.
    \item \texttt{10\_shap\_mechanism.py}--\texttt{13b\_monotonicity\_deep\_dive.py}: Out-of-sample SHAP mechanism tracing and loan-amount decile decomposition.
    \item \texttt{14\_model\_robustness.py}: Logistic regression architectural robustness check.
    \item \texttt{15\_holdout\_validation.py}, \texttt{16\_oos\_mechanism\_shap.py}: Strict holdout replication of headline recall gaps and SHAP attributions.
    \item \texttt{17\_final\_methods\_checks.py}: Formal $z$-tests, chi-square tests, Cram\'er's $V$, Spearman correlations, and coercion-attrition audit.
\end{itemize}

\subsection{Software Dependencies}

Python 3.11 with pandas $\geq$2.1.0, numpy $\geq$1.26.0, scikit-learn $\geq$1.3.0, xgboost $\geq$2.0.0, shap $\geq$0.43.0, statsmodels $\geq$0.14.0, scipy $\geq$1.11.0.

\subsection{Determinism}

A global seed of \texttt{random\_state=42} is applied across all data partitioning functions, imputers, and model initializations. All cross-validation is stratified jointly on default status and income bracket.

\end{document}